%% file: cltransfer.tex
\documentclass[11pt,a4paper]{article}
\pdfoutput=1
\PassOptionsToPackage{breaklinks}{hyperref}
\usepackage[hyperref]{naaclhlt2019}
\usepackage{times}
\usepackage{amssymb}
\usepackage{amsmath}
\usepackage{latexsym}
\usepackage{booktabs}
\usepackage{tabulary}
\usepackage{array}
\newcolumntype{P}[1]{>{\centering\arraybackslash}m{#1}}
\usepackage{graphicx}
\usepackage{xcolor}
\usepackage{soul}
\usepackage{sidecap}

\newcommand{\lang}[1]{\textsc{#1}}
\newcommand{\eng}{\lang{eng}}
\newcommand{\cmn}{\lang{cmn}}
\newcommand{\ara}{\lang{ara}}

\definecolor{darkgreen}{HTML}{228B22}

\newcommand{\squishlist}{
 \begin{list}{$\bullet$}
  { \setlength{\itemsep}{0pt}
     \setlength{\parsep}{1pt}
     \setlength{\topsep}{1pt}
     \setlength{\partopsep}{0pt}
     \setlength{\leftmargin}{1.5em}
     \setlength{\labelwidth}{1em}
     \setlength{\labelsep}{0.5em} } }
 \newcommand{\squishend}{\end{list}}

\aclfinalcopy 
\def\aclpaperid{1875} 

\usepackage{xspace}
\newcommand{\cwr}{\textsc{cwr}\xspace}
\newcommand{\monochar}{\textsc{MonoChar}\xspace}
\newcommand{\polychar}{\textsc{RositaChar}\xspace}
\newcommand{\polyword}{\textsc{RositaWord}\xspace}
\newcommand{\rosita}{Rosita\xspace}

\title{Polyglot Contextual Representations Improve Crosslingual Transfer}

\author{
    Phoebe Mulcaire$^{\heartsuit}$ \quad
	Jungo Kasai$^{\heartsuit}$ \quad Noah A. Smith$^{\heartsuit \diamondsuit}$\\
    $^{\heartsuit}$Paul G.~Allen School of Computer Science \& Engineering,\\University of Washington, Seattle, WA, USA \\
    $^{\diamondsuit}$Allen Institute for Artificial Intelligence, Seattle, WA, USA \\
    {\tt \{pmulc,jkasai,nasmith\}@cs.washington.edu}
}

\date{}

\begin{document}

\maketitle
\begin{abstract}
We introduce \rosita, a method to produce multilingual contextual word representations by training a single language model on text from multiple languages. Our method combines the advantages of contextual word representations with those of multilingual representation learning.
We produce language models from dissimilar language pairs (English/Arabic and English/Chinese) and use them in dependency parsing, semantic role labeling, and named entity recognition, with comparisons to monolingual and non-contextual variants.
Our results provide further evidence for the benefits of polyglot learning, in which representations are shared across multiple languages.
\end{abstract}

\section{Introduction}

State-of-the-art methods for crosslingual transfer make use of multilingual word embeddings, and much research has explored methods that align vector spaces for words in different languages \cite{faruqui2014improving,upadhyay2016cross,ruder2017survey}. On the other hand, 
\emph{contextual} word representations (\cwr) extracted from language models (LMs) have advanced the state of the art beyond what was achieved with word type representations on many monolingual NLP tasks \citep{Peters2018}.
Thus, the  question arises:  can \emph{contextual} word representations benefit from \emph{multilinguality}?

We introduce a method to produce multilingual \cwr by training a single ``polyglot'' language model on text in multiple languages. As our work is a multilingual extension of ELMo \cite{Peters2018}, we call it \rosita (after a bilingual character from \emph{Sesame Street}).
Our hypothesis is that, although each language is unique, different languages manifest similar characteristics (e.g., morphological, lexical, syntactic) which can be exploited by training a single model with data from multiple languages \cite{ammarthesis}.
Previous work has shown this to be true to some degree in the context of syntactic dependency parsing \cite{Ammar2016ManyLO}, semantic role labeling \cite{Mulcaire2018PolyglotSR}, named entity recognition \cite{xie2018ner}, and language modeling for phonetic sequences \cite{Tsvetkov2016} and for speech recognition \cite{Ragni2016MultiLanguageNN}.
Recently, \citet{delhoneux-EtAl:2018:EMNLP} showed that parameter sharing between languages can improve performance in dependency parsing, but the effect is variable, depending on the language pair and the parameter sharing strategy.
Other recent work also reported that concatenating data from different languages can hurt performance in dependency parsing \cite{Che2018ElmoUD}.
These mixed results suggest that while crosslingual transfer in neural network models is a promising direction, the best blend of polyglot and language-specific elements may depend on the task and architecture.
However, we find overall contextual representations from polyglot language models succeed in a range of settings, even where multilingual word type embeddings do not, and are a useful technique for crosslingual transfer.

We explore crosslingual transfer between highly dissimilar languages (English$\rightarrow$Chinese and English$\rightarrow$Arabic) for three core tasks: Universal Dependency (UD) parsing, semantic role labeling (SRL), and named entity recognition (NER).
We provide some of the first work using polyglot LMs to produce contextual representations,\footnote{Contemporaneous work uses   polyglot LMs for natural language inference and machine translation \citep{lample2019}.} and the first analysis comparing them to monolingual LMs for this purpose. We also introduce an LM variant which takes multilingual word embedding input as well as character input, and explore its applicability for producing contextual word representations.
Our experiments focus on comparisons in three dimensions:
monolingual vs.~polyglot representations, contextual vs.~word type embeddings, and, within the contextual representation paradigm, purely character-based language models vs.~ones that include word-level input.

Previous work has shown that contextual representations offer a significant advantage over traditional word embeddings (word type representations). In this work, we show that, on these tasks, polyglot character-based language models can provide benefits on top of those offered by contextualization.
Specifically, even when crosslingual transfer with word type embeddings hurts target language performance relative to monolingual models, polyglot \textit{contextual} representations can improve target language performance relative to monolingual versions, suggesting that polyglot language models tie dissimilar languages in an effective way.

In this paper, we use the following terms: \textit{crosslingual transfer} and \textit{polyglot learning}.
While crosslingual transfer is often used in situations where target data are absent or scarce, we use it broadly to mean any method which uses one or more source languages to help process another target language. 
We also draw a sharp distinction between multilingual and polyglot models.
Multilingual learning can happen independently for different languages, but a polyglot solution  provides a single model for multiple languages, e.g., by parameter sharing between languages in networks during training.

\section{Polyglot Language Models}
We first describe the language models we use to construct multilingual (and monolingual) \cwr.

\subsection{Data and Preprocessing}
Because the Universal Dependencies treebanks we use for the parsing task predominantly use Traditional Chinese characters and the Ontonotes data for SRL and NER consist of Simplified Chinese, we train separate language models for the two variants. For English we use text from the Billion Word Benchmark \cite{chelba2013one}, for Traditional Chinese, wiki and web data provided for the CoNLL 2017 Shared Task \cite{Conll2017data}, for Simplified Chinese, newswire text from Xinhua,\footnote{\url{catalog.ldc.upenn.edu/LDC95T13}} and for Arabic, newswire text from AFP.\footnote{\url{catalog.ldc.upenn.edu/LDC2001T55}} We use approximately 60 million tokens of news and web text for each language.

We tokenized the language model training data for English and Simplified Chinese using Stanford CoreNLP \cite{manning-EtAl:2014:P14-5}. The Traditional Chinese corpus was already pre-segmented by UDPipe \cite{Conll2017data,Straka2016UDPipeTP}. We found that the Arabic vocabulary from AFP matched both the UD and Ontonotes data reasonably well without additional tokenization. We also processed all corpora to normalize punctuation and remove non-text.

\subsection{Models and Training}

We base our language models on the ELMo method \cite{Peters2018}, which encodes each word with a character CNN, then processes the word in context with a word-level LSTM.\footnote{A possible alternative is BERT \cite{Devlin2018}, which uses a  bidirectional objective
and a transformer architecture in place of the LSTM. Notably, one of the provided BERT models was trained on several languages in combination, in a simple polyglot approach (see \url{https://github.com/google-research/bert/blob/master/multilingual.md}). Our initial exploration of multilingual BERT models raised sufficient questions about preprocessing that we defer exploration to future work.
}
Following \newcite{Che2018ElmoUD}, who used 20 million words per language to train monolingual language models for many languages, we use the same hyperparameters used to train the monolingual English language model from \newcite{Peters2018}, except that we reduce the internal LSTM dimension from 4096 to 2048.

For each target language dataset (Traditional Chinese, Simplified Chinese, and Arabic), we produce:
\squishlist
    \item a monolingual language model with character CNN (\monochar) trained on that language's data;
    \item a polyglot LM (\polychar) trained with the same code, on that language's data with an additional, equal amount of English data;
    \item a modified polyglot LM (\polyword), described below. 
\squishend

The \polyword model concatenates a 300 dimensional word type embedding, initialized with multilingual word embeddings, to the character CNN encoding of the word, before passing this combined vector to the bidirectional LSTM. The idea of this word-level initialization is to bias the model toward crosslingual sharing; because words with similar meanings have similar representations, the features that the model learns are expected to be at least partially language-agnostic. The word type embeddings used for these models, as well as elsewhere in the paper, are trained on our language model training set using the fastText method \cite{bojanowski2017enriching}, and target language vectors are aligned with the English ones using supervised MUSE\footnote{For our English/Chinese and English/Arabic data, their unsupervised method yielded substantially worse results in word translation.} \cite{conneau2017word}. 
See appendix for more LM training details.

\section{Experiments}

All of our task models (UD, SRL, and NER) are implemented in AllenNLP, version 0.7.2 \cite{Gardner2017AllenNLP}.\footnote{We make our multilingual fork available at \url{https://github.com/pmulcaire/rosita}}
We generally follow the default hyperparameters and training schemes provided in the AllenNLP library regardless of language. See appendix for the complete list of our hyperparameters.
For each task, we experiment with five types of word representations: in addition to the three language model types (\monochar, \polychar, and \polyword) described above, we show results for the task models trained with monolingual and polyglot non-contextual word embeddings.

After pretraining, the word representations are fine-tuned to the specific task during task training.
In non-contextual cases, we fine-tune by updating word embeddings directly, while in contextual cases, we only update coefficients for a linear combination of the internal representation layers for efficiency \cite{Peters2018}. 
In order to properly evaluate our models' generalization ability, we ensure that  sentences in the test data are excluded from the data used to train the language models.

\subsection{Universal Dependency Parsing}
\begin{SCtable*}
    \begin{tabulary}{\textwidth}{|l|l|c|c|c|}
    \hline
vectors& \shortstack{task lang.} & UD LAS  & SRL $F_1$ & NER $F_1$ \\ \hline
    fastT (\cmn)    & \cmn       & 85.15$_{\pm 0.12}$ &69.79& 76.31  \\
    fastT (\cmn+\eng)& \cmn+\eng & 84.92$_{\pm 0.28}$ &70.82&  76.05  \\
    \monochar      & \cmn       & 87.55$_{\pm 0.25}$  &74.14&  78.18  \\
    \polychar       & \cmn       & 87.16$_{\pm 0.08}$ &74.24&  \textbf{78.29} \\
    \polychar       & \cmn+\eng &\bf 87.75$_{\pm 0.16}$ &74.69&  77.68  \\
    \polyword       & \cmn       & 86.50$_{\pm 0.17}$ &\textbf{74.84}& 77.19   \\
    \polyword       & \cmn+\eng & 86.37$_{\pm 0.35}$ &74.69& 77.16 \\ 
    Best prior work &\cmn&--&62.83&75.63\\
    \hline
    fastT (\ara)    & \ara      & 82.58$_{\pm 0.51}$ &50.50& 71.60 \\
    fastT (\ara+\eng)& \ara+\eng & 82.67$_{\pm 0.46}$ &54.82& 71.45  \\
    \monochar       & \ara      & 84.98$_{\pm 0.18}$ &\textbf{59.55}& 75.02 \\
    \polychar       & \ara      & 84.98$_{\pm 0.12}$ &58.69& 75.56 \\
    \polychar       & \ara+\eng & \bf 85.24$_{\pm 0.13}$ &59.29&  \textbf{76.19} \\
    \polyword       & \ara      & 84.34$_{\pm 0.20}$ &58.34& 74.02\\
    \polyword       & \ara+\eng & 84.24$_{\pm 0.13}$ &59.47&72.79 \\
    Best prior work &\ara&--&48.68&68.02\\\hline
    \end{tabulary}
    \caption{LAS for UD parsing, $F_1$ for SRL, and $F_1$ for NER, with different 
    input representations. For UD, each number is an average over five runs with different initialization, with standard deviation.  
    SRL/NER results are from one run. The ``task lang.'' column indicates whether the UD/SRL/NER model was trained on annotated text in the target language alone, or a blend of English and the target language data.
    \polyword LMs use as word-level input the same multilingual word vectors as fastText models. The best prior result for Ontonotes Chinese NER is in \citet{Shen2017DeepAL}; the others are from \citet{Pradhan2013TowardsRL}.}
    \label{tab:full_comparison}
\end{SCtable*}

\begin{table}
    \begin{center}
    \begin{tabulary}{\textwidth}{|l|l|c|}
    \hline
    LM type  & \shortstack{task lang.} & LAS \\ \hline
    \multicolumn{2}{|c|}{Harbin \citep{Che2018ElmoUD} \cmn}  & 76.77 \\ 
    \multicolumn{2}{|c|}{Harbin (non-ensemble) \cmn}  & 75.55 \\ 
    \polychar  & \cmn            & 77.40\\
    \polychar & \cmn+\eng       &  \bf 77.63\\ 
    \hline
    \multicolumn{2}{|c|}{Stanford \citep{qi-EtAl:2018:K18-2} \ara}& 77.06 \\ 
    \polychar  & \ara            & 77.79 \\
    \polychar  & \ara+\eng       & \bf 78.02\\
    \hline
    \end{tabulary}
    \caption{LAS ($F_1$) comparison to the winning systems for each language in the CoNLL 2018 shared task for UD. We use predicted POS and the segmentation of the winning system for that language. The \polychar LM variant was selected based on development performance in the gold-segmentation condition.}
    \label{tab:ud_sota_narrow}
    \end{center}
\end{table}

We use a state-of-the-art graph-based dependency parser with BiLSTM and biaffine attention \cite{dozatmanning2017}.
Specifically, the parser takes as input word representations and 100-dimensional fine-grained POS embeddings following \citet{dozatmanning2017}.
We use the same UD treebanks and train/dev./test splits as the CoNLL 2018 shared task on multilingual dependency parsing \cite{zeman-EtAl:2018:K18-2}.
In particular, we use the GUM treebank for English,\footnote{While there are several UD English corpora, we choose the GUM corpus to minimize domain mismatch.} GSD for Chinese, and PADT for Arabic.
For training and validation, we use the provided gold POS tags and word segmentation.

For each configuration, we run experiments five times with random initializations and report the mean and standard deviation.
For testing, we use the CoNLL 2018 evaluation script and consider two scenarios: (1) gold POS tags and word segmentations and (2) predicted POS tags and word segmentations from the system outputs of \citet{Che2018ElmoUD} and \citet{qi-EtAl:2018:K18-2}.\footnote{System outputs for all systems are available at \url{https://lindat.mff.cuni.cz/repository/xmlui/handle/11234/1-2885}}
The former scenario enables us to purely assess parsing performance; see column 3 in Table \ref{tab:full_comparison} for these results on Chinese and Arabic.
The latter allows for a direct comparison to the best previously reported parsers (Chinese, \citealp{Che2018ElmoUD}; Arabic, \citealp{qi-EtAl:2018:K18-2}). See Table \ref{tab:ud_sota_narrow} for these results.

As seen in Table \ref{tab:full_comparison}, the Universal Dependencies results generally show a significant improvement from the use of \cwr.
The best results for both languages come from the \polychar LM and polyglot task models, showing that polyglot training helps, but that the word-embedding initialization of the \polyword model does not necessarily lead to a better final model.
The results also suggest that combining \polychar LM and polyglot task training is key to improve parsing performance.
Table \ref{tab:ud_sota_narrow} shows that we outperform the state-of-the-art systems from the shared task competition. In particular, our LMs even outperform the Harbin system, which uses monolingual \cwr and an ensemble of three biaffine parsers.

\subsection{Semantic Role Labeling}

We use a strong existing model based on BIO tagging on top of a deep interleaving BiLSTM with highway connections \cite{He2017DeepSR}.
The SRL model takes as input word representations and 100-dimensional predicate indicator embeddings following \citet{He2017DeepSR}.
We use a standard PropBank-style, span-based SRL dataset for English, Chinese, and Arabic: Ontonotes \cite{Pradhan2013TowardsRL}.
Note that Ontonotes provides annotations using a single shared annotation scheme for English, Chinese, and Arabic, which can facilitate crosslingual transfer.
For Chinese and English we simply use the provided surface form of the words. The Arabic text in Ontonotes has diacritics to indicate vocalization which do not appear (or only infrequently) in the original source or in our language modeling data. We remove these for better consistency with the language model vocabulary. 
We use gold predicates and the CoNLL 2005 evaluation script for the experiments below to ensure our results are comparable to prior work. See column 4 in Table \ref{tab:full_comparison} for results on the CoNLL-2012 Chinese and Arabic test sets.

The SRL results confirm the advantage of \cwr. Unlike the other two tasks, multilingual word type embeddings are better than monolingual versions in SRL. Perhaps relatedly, models using \polyword are more successful here, providing the highest performance on Chinese. One unusual result is that the model using the \monochar LM is most successful for Arabic. 
This may be linked to the poor results on Arabic SRL overall, which are likely due to the much smaller size of the corpus compared to Chinese (less than 20\% as many annotated predicates) and higher proportion of language-specific tags. 
Such language-specific tags in Arabic could limit the effectiveness of shared English-Arabic representations.
Still, polyglot methods' performance is only slightly behind.

\subsection{Named Entity Recognition}

We use the state-of-the-art BiLSTM-CRF NER model with the BIO tagging scheme \cite{Peters2017SemisupervisedST}.
The network takes as input word representations and 128-dimensional character-level embeddings from a character LSTM. 
We again use the Ontonotes dataset with the standard data splits.
See the last column in Table \ref{tab:full_comparison} for results on the CoNLL-2012 Chinese and Arabic test sets. As with most other experiments, the NER results show a strong advantage from the use of contextual representations and a smaller additional advantage from those produced by polyglot LMs.

\section{Discussion}
Overall, our results show that polyglot language models produce very useful representations.
While Universal Dependency parsing, Arabic SRL, and Chinese NER show models using contextual representations outperform those using word type representations, the advantage from polyglot training in some cases is minor.
However, Chinese SRL and Arabic NER show strong improvement both from contextual word representations and from polyglot training.
Thus, while the benefit of crosslingual transfer appears to be somewhat variable and task dependent, polyglot training is helpful overall for contextual word representations. 
Notably, the \polychar LM does not involve any direct supervision of tying two languages together, such as bilingual dictionaries or parallel corpora, yet is still most often able to learn the most effective representations. One explanation is that it automatically learns crosslingual connections from unlabeled data alone. Another possibility, though, is that the additional data provided in polyglot training produces a useful regularization effect, improving the target language representations without crosslingual sharing (except that induced by shared vocabulary, e.g., borrowings, numbers, or punctuation). 
Nevertheless, the success of polyglot language models is worth further study.

\section{Conclusion}
We presented a method for using polyglot language models to produce multilingual, contextual word representations, and demonstrated their benefits, producing state-of-the-art results in multiple tasks. These results provide a foundation for further study of polyglot language models and their use as unsupervised components of multilingual models.

\section*{Acknowledgments}
The authors thank Mark Neumann for assistance with the AllenNLP library and the anonymous reviewers for their helpful feedback.
This research was funded in part by NSF grant IIS-1562364, the Funai Overseas Scholarship to JK, and the NVIDIA Corporation through the donation of a GeForce GPU.

\bibliography{cltransfer}
\bibliographystyle{acl_natbib}


\clearpage
\input{appendix}

\end{document}

%% file: appendix.tex
\appendix
\newpage

\section{Supplementary Material}
In this supplementary material, we provide hyperparameters used in our models for easy replication of our results.

\subsection{Language Models}
\begin{table}
\small
\centering
\begin{tabular}{ |l p{0.3\linewidth}|}
\hline
\multicolumn{2}{|c|}{Character CNNs}\\
Char embedding size & 16\\
(\# Window Size, \# Filters) & (1, 32), (2, 32), (3, 68), (4, 128), (5, 256), 6, 512), (7, 1024)\\ 
Activation & Relu\\
\hline
\multicolumn{2}{|c|}{Word-level LSTM}\\
LSTM size & 2048\\
\# LSTM layers & 2\\
LSTM projection size & 256\\
Use skip connections & Yes\\
Inter-layer dropout rate& 0.1\\
\hline
\multicolumn{2}{|c|}{Training}\\
Batch size & 128\\
Unroll steps (Window Size) & 20\\
\# Negative samples & 64\\
\# Epochs & 10\\
Adagrad \cite{journals/jmlr/DuchiHS11} lrate& 0.2\\
Adagrad initial accumulator value& 1.0 \\
\hline
\end{tabular}
\caption{Language Model Hyperparameters.}
\label{lm-hyp}
\end{table}

Seen in Table \ref{lm-hyp} is a list of hyperparameters for our language models. 
We generally follow \citet{Peters2018} and use their publicly available code for training.\footnote{\url{https://github.com/allenai/bilm-tf}} 
For character only models, we halve the LSTM and projection sizes to expedite training and to compensate for the greatly reduced training data---their hyperparameters were tuned on around 30M sentences, while we used less than 3M sentences (60-70M tokens) per language.

\subsection{UD Parsing}
\begin{table}
\small
\centering
\begin{tabular}{ |l l|}
\hline
\multicolumn{2}{|c|}{Input}\\
POS embedding size & 100\\
Input dropout rate & 0.3\\
\hline
\multicolumn{2}{|c|}{Word-level BiLSTM}\\
LSTM size & 400\\
\# LSTM layers & 3\\
Recurrent dropout rate & 0.3\\
Inter-layer dropout rate & 0.3\\
Use Highway Connection& Yes\\
\hline
\multicolumn{2}{|c|}{Multilayer Perceptron, Attention}\\
Arc MLP size & 500\\ 
Label MLP size & 100\\ 
\# MLP layers & 1\\
Activation & Relu\\
\hline
\multicolumn{2}{|c|}{Training}\\
Batch size & 80\\
\# Epochs & 80\\
Early stopping & 50\\
Adam \cite{Kingma2015} lrate& 0.001\\
Adam $\beta_1$& 0.9\\
Adam $\beta_2$& 0.999\\
\hline
\end{tabular}
\caption{UD Parsing Hyperparameters.}
\label{ud-hyp}
\end{table}
For UD parsing, we generally follow the hyperparameters provided in AllenNLP \cite{Gardner2017AllenNLP}. See a list of hyperparameters in Table \ref{ud-hyp}.

\subsection{Semantic Role Labeling}
\begin{table}
\small
\centering
\begin{tabular}{ |l l|}
\hline
\multicolumn{2}{|c|}{Input}\\
Predicate indicator embedding size & 100\\
\hline
\multicolumn{2}{|c|}{Word-level Alternating BiLSTM}\\
LSTM size & 300\\
\# LSTM layers & 4\\
Recurrent dropout rate & 0.1\\
Use Highway Connection& Yes\\
\hline
\multicolumn{2}{|c|}{Training}\\
Batch size & 80\\
\# Epochs & 80\\
Early stopping & 20\\
Adadelta \cite{Zeiler2012ADADELTAAA} lrate& 0.1\\
Adadelta $\rho$ & 0.95\\
Gradient clipping & 1.0\\
\hline
\end{tabular}
\caption{SRL Hyperparameters.}
\label{srl-hyp}
\end{table}
For SRL, we again follow the hyperparameters given in AllenNLP (Table \ref{srl-hyp}).
The one exception is that we used 4 layers of alternating BiLSTMs instead of 8 layers to expedite the training process. 

\subsection{Named Entity Recognition}
\begin{table}
\small
\centering
\begin{tabular}{ |l l|}
\hline
\multicolumn{2}{|c|}{Char-level LSTM}\\
Char embedding size & 25\\
Input dropout rate & 0.5\\
LSTM size & 128\\
\# LSTM layers & 1\\
\hline
\multicolumn{2}{|c|}{Word-level BiLSTM}\\
LSTM size & 200\\
\# LSTM layers & 3\\
Inter-layer dropout rate & 0.5\\
Recurrent dropout rate & 0.5\\
Use Highway Connection & Yes\\
\hline
\multicolumn{2}{|c|}{Multilayer Perceptron}\\
MLP size & 400\\ 
Activation & tanh\\
\hline
\multicolumn{2}{|c|}{Training}\\
Batch size & 64\\
\# Epochs & 50\\
Early stopping & 25\\
Adam \cite{Kingma2015} lrate& 0.001\\
Adam $\beta_1$& 0.9\\
Adam $\beta_2$& 0.999\\
L2 regularization coefficient&0.001 \\
\hline
\end{tabular}
\caption{NER Hyperparameters.}
\label{ner-hyp}
\end{table}
We again use the hyperparameter configurations provided in AllenNLP. See Table \ref{ner-hyp} for details.